\documentclass[journal]{IEEEtran}
\usepackage{amsmath,amsfonts}
\usepackage{algorithmic}
\usepackage{array}
\usepackage[caption=false,font=normalsize,labelfont=sf,textfont=sf]{subfig}
\usepackage{textcomp}
\usepackage{stfloats}
\usepackage{url}
\usepackage{verbatim}
\usepackage{graphicx}

\usepackage[numbers,sort&compress]{natbib}
\usepackage[table]{xcolor} 
\usepackage{booktabs}
\usepackage{multirow}
\usepackage{balance}
\usepackage[pagebackref=true,breaklinks=true,letterpaper=true,colorlinks,bookmarks=false]{hyperref}

\def\BibTeX{{\rm B\kern-.05em{\sc i\kern-.025em b}\kern-.08em
    T\kern-.1667em\lower.7ex\hbox{E}\kern-.125emX}}

\begin{document}
\title{Vision Transformers for Single Image Dehazing}

\author{
Yuda Song, Zhuqing He, Hui Qian, and Xin Du.
\thanks{
Manuscript received XXXX 00, 0000; accepted XXXX 00, 0000. 
Date of publication XXXX 00, 0000; date of current version XXXX 00, 0000. 
The associate editor coordinating the review of this manuscript and approving it for publication was XXXX.
(Yuda Song and Zhuqing He contributed equally to this work.) (Corresponding authors: Xin Du.)
 
Yuda Song, Zhuqing He, Hui Qian, and Xin Du are with Zhejiang University, Hangzhou 310027, China
(e-mail: duxin@zju.edu.cn)
}
}

\markboth{Journal of \LaTeX\ Class Files,~Vol.~18, No.~9, September~2020}%
{How to Use the IEEEtran \LaTeX \ Templates}

\maketitle

\newcommand{\gr}{\rowcolor[gray]{.95}}
\newcommand{\rt}{\textcolor[rgb]{0.75,0.25,0.25}}
\newcommand{\bt}{\textcolor[rgb]{0.25,0.25,0.75}}

\begin{abstract}
    Image dehazing is a representative low-level vision task that estimates latent haze-free images from hazy images.
    In recent years, convolutional neural network-based methods have dominated image dehazing.
    However, vision Transformers, which has recently made a breakthrough in high-level vision tasks, has not brought new dimensions to image dehazing.
    We start with the popular Swin Transformer and find that several of its key designs are unsuitable for image dehazing.
    To this end, we propose DehazeFormer, which consists of various improvements, such as the modified normalization layer, activation function, and spatial information aggregation scheme.
    We train multiple variants of DehazeFormer on various datasets to demonstrate its effectiveness.
    Specifically, on the most frequently used SOTS indoor set, our small model outperforms FFA-Net with only 25\% \#Param and 5\% computational cost.
    To the best of our knowledge, our large model is the first method with the PSNR over 40 dB on the SOTS indoor set, dramatically outperforming the previous state-of-the-art methods.
    We also collect a large-scale realistic remote sensing dehazing dataset for evaluating the method's capability to remove highly non-homogeneous haze.
    We share our code and dataset at \href{https://github.com/IDKiro/DehazeFormer}{https://github.com/IDKiro/DehazeFormer}.
\end{abstract}



\section{Introduction}

\IEEEPARstart{H}{aze}
 is a common atmospheric phenomenon that can impair daily life and machine vision systems.
The presence of haze reduces the scene's visibility and affects people's judgment of the object, and thick haze can even affect traffic safety.
For computer vision, haze degrades the quality of the captured image in most cases. 
It can impact the model's reliability in high-level vision tasks, further mislead machine systems, such as autonomous driving.
All these make image dehazing a meaningful low-level vision task.

Image dehazing aims to estimate the latent haze-free image from the observed hazy image.
For the single image dehazing problem, there is a popular model~\cite{mccartney1976optics,nayar1999vision,narasimhan2002vision} to characterize the degradation process for hazy images:
\begin{equation}
    \label{eq:haze}
    I = J(x) t(x) + A (1 - t(x)),
\end{equation}
where $I$ is the captured hazy image, $J$ is the latent haze-free image, $A$ is the global atmospheric light, and $t$ is the medium transmission map.
And the transmission can be expressed as 
\begin{equation}
    \label{eq:depth}
    t(x) = e^{-\beta d(x)},
\end{equation}
where $\beta$ is the scattering coefficient of the atmosphere, and $d$ is the scene depth.
As can be seen, image dehazing is a typically ill-posed problem, and early image dehazing methods tend to constrain the solution space with priors~\cite{he2010single,fattal2014dehazing,zhu2015fast,berman2016non}.
They generally estimate $A$ and $t(x)$ separately to lower the complexity of the problem and then use Eq.(\ref{eq:haze}) to derive the results.
These prior-based methods can produce images with good visibility. 
However, these images are often visibly different from haze-free images, and artifacts may be introduced in regions that do not satisfy the priors.

In recent years, deep learning has made a big hit in computer vision, and researchers have proposed a large number of image dehazing methods based on deep convolutional neural networks (CNNs)~\cite{cai2016dehazenet,ren2016single,zhang2018densely,li2017aod,ren2018gated,liu2019griddehazenet,chen2019gated,dong2020physics,deng2020hardgan,dong2020multi,qin2020ffa,wu2021contrastive,wang2021eaa,shao2020domain}.
With a sufficient number of synthetic image pairs, these methods can achieve superior performance over prior-based methods.
Earlier CNN-based methods~\cite{cai2016dehazenet,ren2016single,zhang2018densely} also estimate $A$ and $t(x)$ separately, where $t(x)$ is supervised using the transmission map used in synthesizing the dataset.
And current methods~\cite{liu2019griddehazenet,chen2019gated,dong2020physics,deng2020hardgan,qin2020ffa,dong2020multi,wu2021contrastive,wang2021eaa} prefer to predict the latent haze-free image or the residuals of the haze-free image versus the hazy image since it tends to achieve better performance.
Very recently, ViT~\cite{vaswani2017attention} outperformed almost all CNN architectures in high-level vision tasks using plain Transformer architecture.
Subsequently, many modified architectures~\cite{wang2021pyramid,yuan2021tokens,choromanski2020rethinking,xiao2021early,dai2021coatnet,li2022uniformer,park2022vision,liu2021swin,huang2021shuffle,yu2021glance,wang2021crossformer,lin2021cat,chen2021regionvit,chen2021crossvit,chu2021twins,dong2021cswin,wu2021pale,yang2021focal} have been proposed, and vision Transformer is challenging the dominance of CNNs in high-level vision tasks.
So many works have demonstrated the effectiveness of vision Transformers, but there is still no Transformer-based image dehazing method that defeats the state-of-the-art image dehazing networks.
In this work, we propose an image dehazing Transformer dubbed DehazeFormer, which is inspired by Swin Transformer~\cite{liu2021swin}.
It dramatically surpasses these CNN-based methods.

\begin{figure}[t]
    \centering
    \includegraphics[width=1.0\columnwidth]{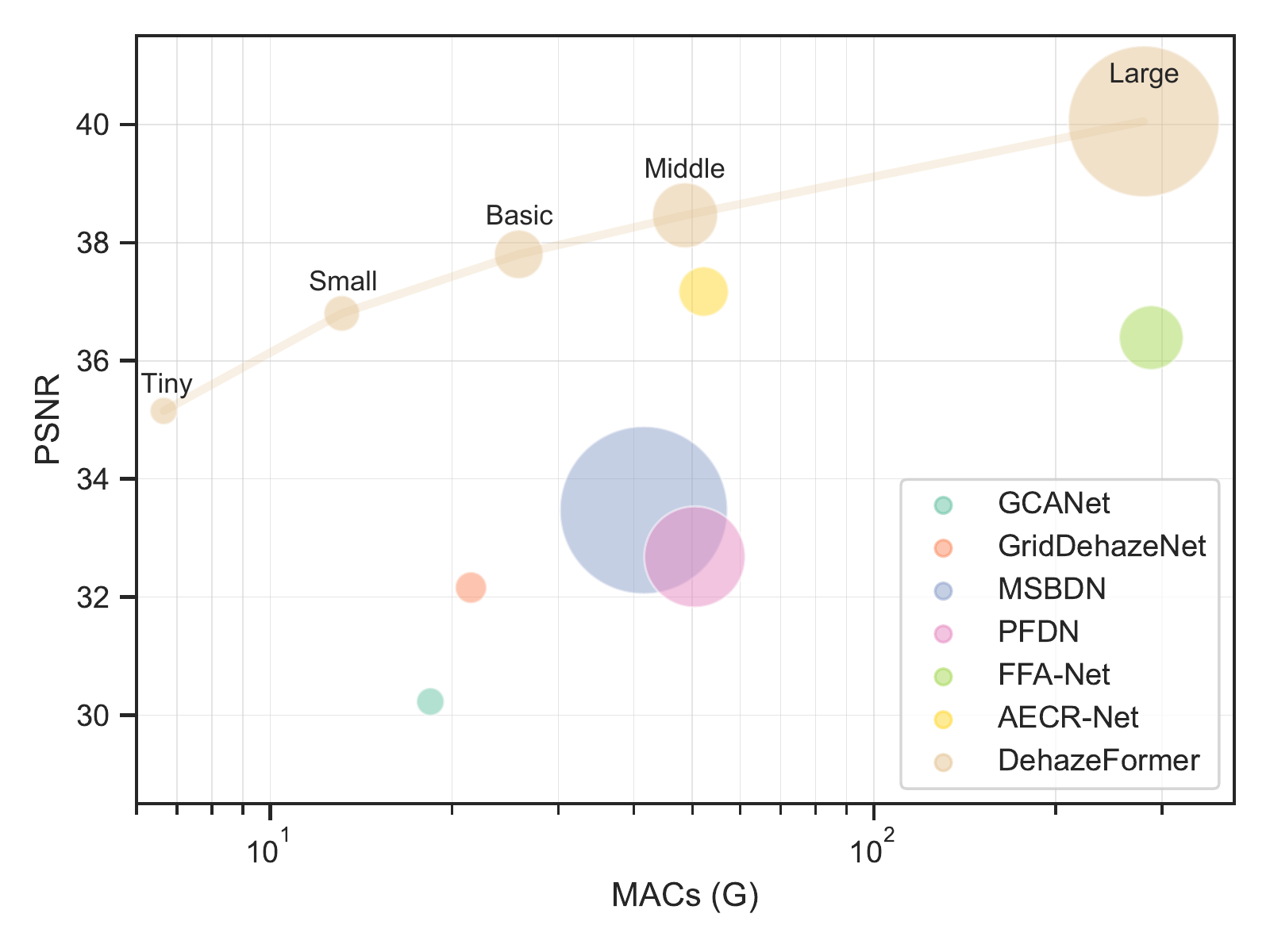}
    \caption{
        Comparison of DehazeFormer with other image dehazing methods on the SOTS indoor set.
        The size of the dots indicates the \#Param of the method, and MACs are shown with logarithmic axis.
    }
    \label{fig:intro}
\end{figure}

We find that the LayerNorm~\cite{ba2016layer} and GELU~\cite{hendrycks2016gaussian} commonly used in vision Transformers harm the image dehazing performance.
Specifically, the LayerNorm used in vision Transformer normalizes the tokens corresponding to the image patches separately, resulting in the loss of the relativity between the patches.
Hence, we remove the normalization layer preceded by the multi-layer perceptron (MLP) and propose RescaleNorm to replace LayerNorm.
RescaleNorm performs normalization on the entire feature map and reintroduces the mean and variance of the feature map lost after normalization.
Besides, SiLU / Swish~\cite{ramachandran2017searching} and GELU work well in high-level vision tasks, but ReLU~\cite{nair2010rectified} works better than them in image dehazing.
We believe this is because the nonlinearities they introduce are not easily inverted when decoding.
We argue that image dehazing requires not only that the network encodes highly expressive features but also that these features are easily recovered to image domain signals.

Swin Transformer uses window partitioning with cyclic shift to efficiently aggregate local features.
But we find that the cyclic shift is suboptimal for image edge regions in image dehazing.
Specifically, the cyclic shift should use masked multi-head self-attention (MHSA) to prevent unreasonable spatial aggregation, making the windows in the edge regions smaller.
We consider that aggregating information within a small window brings instability, which can bias the network's training.
Thus we propose a shifted window partitioning scheme based on reflection padding and cropping, which allows MHSA to discard the mask and achieve a constant window size.
We also find that the aggregation weights of MHSA are always positive, which makes it behave like a low-pass filter~\cite{park2022vision}.
Because the aggregation weights of MHSA are dynamic, all-positive, and normalized, we believe that static, learnable, and unconstrained aggregation weights are helpful to complement the MHSA, while the convolution meets this criterion.

Furthermore, we propose a prior-based soft reconstruction module that outperforms global residual learning and a multi-scale feature map fusion module based on SKNet~\cite{li2019selective} to replace concatenation fusion.
Finally, we build multiple U-Net-like image dehazing Transformers using the proposed modules.
Our experiments show that DehazeFormer can substantially outperform contemporaneous methods with lower overhead.
Fig.~\ref{fig:intro} shows the comparison of DehazeFormer with other image dehazing methods on the SOTS indoor set.
Our small model defeats the FFA-Net~\cite{qin2020ffa} with only 25\% \#Param and 5\% computational cost.
Our base model is lower in overhead but better in performance than the previous state-of-the-art method, AECR-Net~\cite{wu2021contrastive}. 
To the best of our knowledge, our large model is the first method over 40 dB, substantially outperforming contemporaneous methods.

There are non-homogeneous image dehazing datasets collected using professional haze machines~\cite{ancuti2020nh}, but they are too small and far from the non-homogeneous haze that would be present in natural scenes.
Instead, we tend to collect remote sensing image dehazing datasets since highly non-homogeneous haze is prevalent in remote sensing images.
We take into account the wavelength, \emph{etc.}, on the spatial distribution of the haze and then synthesize a large-scale realistic remote sensing image dehazing dataset.



\section{Related Works}

\subsection{Image Dehazing}

Early single-image dehazing methods were generally based on the handcraft priors, such as dark channel prior (DCP)~\cite{he2010single}, color attenuation prior (CAP)~\cite{zhu2015fast}, color-lines~\cite{fattal2014dehazing}, and haze-lines~\cite{berman2016non}.
These prior-based methods usually yield images with good visibility.
However, because these priors are based on empirical statistics, these dehazing methods tend to output unrealistic results when the scenes do not satisfy these priors.
With the rapid development of deep learning, learning-based dehazing methods have dominated in recent years.
DehazeNet~\cite{cai2016dehazenet} and MSCNN~\cite{ren2016single} are the pioneers in applying CNNs for image dehazing.
They learn to estimate $t$ and obtain the result together with $A$ estimated by the conventional method.
After that, DCPDN~\cite{zhang2018densely} uses two sub-networks to estimate $t$ and $A$ respectively, while GFN~\cite{ren2018gated} estimates the fusion coefficient maps for the three predefined image operations.
AOD-Net~\cite{li2017aod}, on the other hand, rewrites Eq.(\ref{eq:haze}) so that the network needs to estimate only one component.
GridDehazeNet~\cite{liu2019griddehazenet} proposes that learning to restore the image is better than estimating $t$, because the latter will fall into suboptimal solutions.
And most recent works~\cite{chen2019gated,dong2020physics,deng2020hardgan,qin2020ffa,dong2020multi,wu2021contrastive,wang2021eaa} tend to estimate the haze-free image or the residual between the haze-free image and the hazy image.

Since the dehazing performance of the learning-based methods dramatically depends on the quality and size of the dataset, several datasets have been proposed.
These dehazing datasets are divided into two main categories: real datasets~\cite{ancuti2018haze,ancuti2018haze2,ancuti2019dense,ancuti2020nh} and synthetic datasets~\cite{ancuti2016d,zhang2017hazerd,li2018benchmarking}.
Real datasets use real haze produced by professional haze machines to generate real hazy images.
Synthetic datasets generally use Eq.(\ref{eq:haze}) to synthesize the corresponding hazy images with haze-free images and depth maps.
Although the real datasets seem to be more attractive, it is difficult to obtain enough image pairs, and the distribution of the haze produced by the haze machine still differs significantly from the real haze. 
Hence, most methods tend to use synthetic datasets for training and testing.
In contrast to these datasets, this paper presents a new synthetic remote sensing image dehazing dataset named RS-Haze for evaluating the method's capability to remove highly non-homogeneous haze.
RS-Haze is larger and more realistic than previous datasets~\cite{qin2018dehazing,guo2020rsdehazenet,huang2020single,mehta2021domain}, taking into account sensor characteristics, haze distribution and particle size, wavelengths of light, and other factors that are overlooked.

\subsection{Vision Transformers}

CNN has dominated most computer vision tasks for years, while recently, the Vision Transformer (ViT)~\cite{dosovitskiy2020image} architectures show the capability of replacing CNNs.
ViT pioneered the direct application of the Transformer architecture~\cite{vaswani2017attention}, which projects images into token sequences via patch-wise linear embedding.
The shortcomings of the original ViT are its weak inductive bias and the quadratic computational cost.
To this end, PVT~\cite{wang2021pyramid} uses the pyramid architecture to introduce multi-scale inductive bias and downsamples the key and value to reduce the computational cost.
T2T-ViT~\cite{yuan2021tokens} uses the unfolding operation just like CNNs for tokenization, and it uses the Performer~\cite{choromanski2020rethinking} to lower the computational cost.
Besides, some works~\cite{xiao2021early,dai2021coatnet,li2022uniformer,park2022vision} employ convolution in the early stages to introduce the inductive bias.
Swin Transformer~\cite{liu2021swin} partitions tokens into windows and performs self-attention within a window to keep the linear computational cost.
It employs the cyclic shift scheme to bridge windows so that adjacent blocks adopt different window partitions.
Since then, many follow-ups to Swin Transformer have been proposed.
For example, some methods bridge windows by reshaping the tensor~\cite{huang2021shuffle,yu2021glance,wang2021crossformer,lin2021cat}; while some methods bridge windows by using tokens with global receptive fields as proxies~\cite{chen2021regionvit,chen2021crossvit,chu2021twins}; and others use modified window partitioning schemes~\cite{dong2021cswin,wu2021pale,yang2021focal}.
Our DehazeFormer can be considered as a combination of Swin Transformer and U-Net~\cite{ronneberger2015u}, but with several critical modifications for image dehazing.

There are also some variants of Swin Transformer for low-level vision tasks.
SwinIR~\cite{liang2021swinir} is one of the pioneers to employ Swin Transformer in low-level vision tasks, which builds a large residual block consisting of stacked Swin Transformer layers and a subsequent convolutional layer.
Uformer~\cite{wang2022uformer} uses Swin Transformer blocks to build a U-Net-like network and inserted depth-wise convolution (DWConv)~\cite{chollet2017xception} in the feed-forward network (FFN) like LocalViT~\cite{li2021localvit}.
However, we found that they perform very poorly in image dehazing.
We attribute this to the fact that they inherit the normalization layer, window partitioning scheme, and activation function from the original Swin Transformer.
There are a few ViT-based dehazing networks proposed, such as HyLoG-ViT~\cite{zhao2021hybrid} and TransWeather~\cite{valanarasu2021transweather}.
However, HyLoG-ViT does not show convincing performance, while TransWeather aims to use a DETR-like framework~\cite{carion2020end} to handle multiple weather conditions simultaneously.

\begin{figure*}[t]
    \centering
    \includegraphics[width=1.0\textwidth]{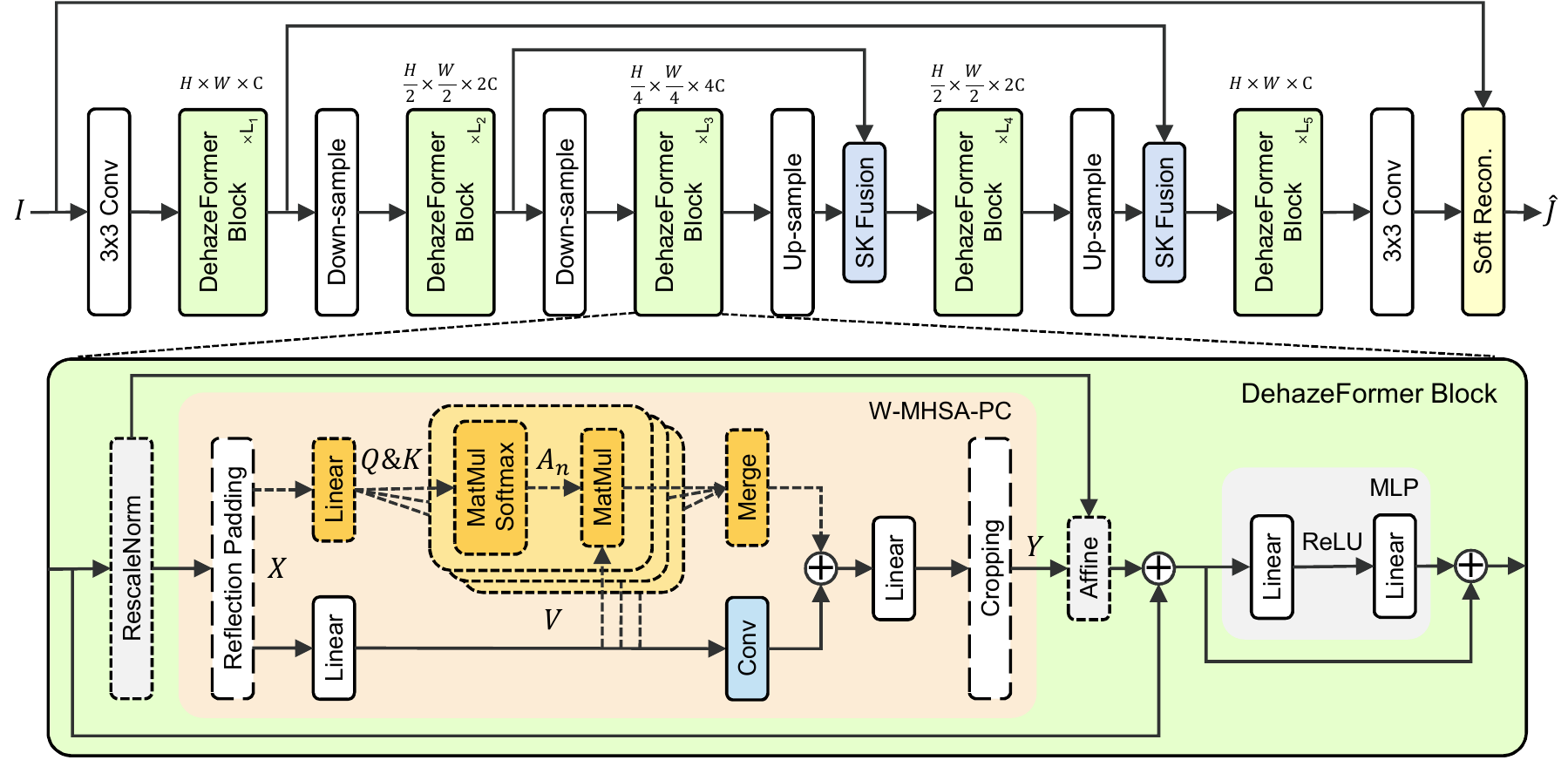}
    \caption{
        DehazeFormer is a modified 5-stage U-Net, whose convolutional blocks are replaced by our DehazeFormer blocks.
        The components illustrated with dashed boxes in the DehazeFormer block indicate they are optional.
        The SK fusion and soft reconstruction layers are proposed to replace the original concatenation fusion and global residual.
        The input size is $H \times W$, and the size of feature maps in each stage is shown below the DehazeFormer block.
    }
    \label{fig:arch}
\end{figure*}

\section{DehazeFormer}

\subsection{Overall}

DehazeFormer's network architecture is based on the popular Swin Transformer~\cite{liu2021swin}, but incorporates several improvements to compensate for the deficiencies of the original Swin Transformer when dealing with image dehazing.
Fig.~\ref{fig:arch} shows the overall architecture of the DehazeFormer.
Given the image pair $\{ I(x), J(x) \}$, we only compute the $L_1$ loss to train the DehazeFormers.

First, we briefly review the Swin Transformer.
Given an input feature map $X \in \mathbb{R} ^ {b \times h \times w \times c}$, we project $X$ to $Q,K,V$ (query, key, value) using linear layers and group tokens using window partitioning.
Swin Transformer applies MHSA within the window, and the window partitioning of adjacent blocks is different.
For simplicity, the following $Q,K,V \in \mathbb{R} ^ {b \times l \times d}$ correspond to a single window \& header, where $l$ is the tokens number in a window and $d$ is the dimension.
Thus the self attention is computed by
\begin{equation}
    \text { Attention }(Q, K, V)=\operatorname{Softmax}\left({Q K^{T}} / \sqrt{d}+B\right) V,
\end{equation}
where $B$ is the relative position bias term.
A linear layer follows it to project the output of the attention.

Our proposed DehazeFormer block differs from the original Swin Transformer block in the normalization layer, the nonlinear activation function, and the spatial information aggregation scheme, detailed in the subsequent subsections.
Besides the DehazeFormer block, the SK fusion and soft reconstruction layers are proposed to replace the concatenation fusion layer and global residual learning.

The SK fusion layer is inspired by SKNet~\cite{li2019selective}, it is designed to fuse multiple branches using channel attention.
Let the two feature maps be $x_1$ and $x_2$, we first use a linear layer $f(\cdot)$ to project $x_1$ to $\hat{x}_1$.
We use the global average pooling $\operatorname{GAP}(\cdot)$, MLP (Linear-ReLU-Linear) $\mathcal{F}_{MLP} (\cdot)$, softmax function and split operation to obtain the fusion weights:
\begin{equation}
    \{a_1, a_2\}=\operatorname{Split}(\operatorname{Softmax}(\mathcal{F}_{MLP}(\operatorname{GAP}\left(\hat{x}_1 + x_2 \right))).
\end{equation}
We use the weights $\{a_1, a_2\}$ to fuse $\hat{x}_1, x_2$ with an additional short residual via $y = a_1 \hat{x}_1 + a_2 x_2 + x_2$.

Current image dehazing networks generally predict the reconstructed image $\hat{J}(x)$ or global residual $R(x)=\hat{J}(x) - I(x)$.
We consider it beneficial to introduce priors, provided that there are no strong constraints since the degradation model is an approximation.
We rewrite Eq.(\ref{eq:haze}) as
\begin{equation}
    \label{eq:dehaze}
    J(x) = K(x) I(x) + B(x) + I(x),
\end{equation}
where $K(x)= {1/t(x) - 1}$ and $B(x)= - \left( {1/t(x) - 1} \right) A$.
We drive the network to predict $O\in \mathbb{R} ^ {h \times w \times 4}$, and split $O$ into $K\in \mathbb{R} ^ {h \times w \times 1}$ and $B\in \mathbb{R} ^ {h \times w \times 3}$.
As a result, the network architecture softly constrains the relationship between $K(x)$ and $B(x)$.
This weak prior allows the network to degenerate to predict global residuals (\emph{i.e.}, $K(x)=\mathbf{0}$, $B(x)=R(x)$).
For convenience, we refer to Eq.(\ref{eq:dehaze}) as soft reconstruction.

\subsection{Rescale Layer Normalization}

The normalization layer plays a vital role in neural network architecture since it stabilizes the network's training.
However, we find that LayerNorm~\cite{ba2016layer}, which Transformers commonly use, may be unsuitable for image dehazing.

We first review the formulation of LayerNorm used by Transformers.
Assume that the shape of the feature map $x \in \mathbb{R} ^ {b \times n \times c}$, where $n = h \times w$ (\emph{i.e.}, height and width), the normalization process can be expressed as:
\begin{equation}
    \hat{x}_{i}=\frac{x_{i}-\mu_{i}}{\sigma_{i}} \cdot \gamma_{i} + \beta_{i}.
    \label{eq:norm}
\end{equation}
Here $\mu$ and $\sigma$ denote the mean and standard deviation, $\gamma$ and $\beta$ are learned scaling factor and bias, and $i=(i_b,i_n,i_c)$ denotes the index.
In LayerNorm, $\mu$ and $\sigma$ are computed along the $c$-axis, making $\mu, \sigma \in \mathbb{R} ^ {b \times n}$.
We believe that the mean and standard deviation are correlated with brightness and contrast for images, so the relative brightness and contrast between image patches are somehow discarded after LayerNorm.
To this ned, we compute $\mu$ and $\sigma$ along the $(n,c)$-axes, leading to $\mu, \sigma \in \mathbb{R} ^ {b}$.
We note this normalization method is the LayerNorm more commonly used in CNNs, referred to as LayerNorm$^\dag$ in this paper.

\begin{figure}[t]
    \centering
    \includegraphics[width=1.0\columnwidth]{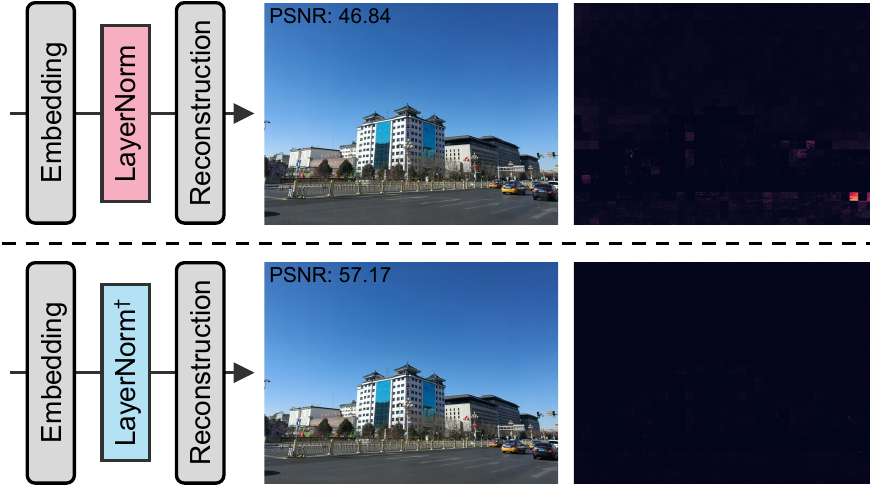}
    \caption{
        Simple autoencoders for analyzing the normalization methods.
        From left to right, there are the autoencoders' architectures, output images, and error maps, where the error is scaled by $8\times$ for better viewing.
        The embedding layer and reconstruction layer are linear layers with patch-wise tensor reshaping.
    }
    \label{fig:rln}
\end{figure}

We conduct a simple experiment to show the negative effects of LayerNorm as shown in Fig.~\ref{fig:rln}.
Specifically, we build autoencoders using only patch embedding, normalization, and patch reconstruction layers.
We train these autoencoders to reconstruct a single input image.
Without global residuals, learning identity mappings is not a trivial task~\cite{he2016deep}.
When LayerNorm is inserted, we can clearly see the block artifacts appearing in the reconstructed image.
Because this autoencoder does not involve interactions between patches, it can only memorize the statistics of the sky region at the expense of the rich-texture region.
By changing LayerNorm to LayerNorm$^\dag$, we largely overcome its negative effects.

However, LayerNorm$^\dag$ still discards the mean and standard deviation of the feature map.
So we propose the Rescale Layer Normalization (RescaleNorm), which is built based on LayerNorm$^\dag$, but the mean and standard deviation computed are saved and introduced at the end of the residual block.
Specifically, we first fetch the $\mu, \sigma \in \mathbb{R} ^ {b \times 1 \times 1}$, and normalize the input feature map $x$ to $\hat{x}$ via Eq.(\ref{eq:norm}).
We then use the main block $\mathbf{F} (\cdot)$ to process $\hat{x}$ to obtain the output $\hat{y}$.
We use two linear layers with weights $W_\gamma, W_\beta \in \mathbb{R} ^ {1 \times c}$ and biases $B_\gamma, B_\beta \in \mathbb{R} ^ {1 \times 1 \times c}$ to transform $\mu$ and $\sigma$ via $\{\gamma, \beta\} = \{\sigma W_\gamma + B_\gamma,\mu W_\beta + B_\beta\}$, where $\gamma, \beta \in \mathbb{R} ^ {b \times 1 \times c}$.
To accelerate convergence, we initialize  $B_\gamma$ and $B_\beta$ to $\mathbf{1}$ and $\mathbf{0}$.
We inject $\gamma$ and $\beta$ into $\hat{y}$ to reintroduce the mean and standard deviation.
Therefore, RescaleNorm can be formulated as:
\begin{equation}
    y=\mathbf{F}  \left ({\frac{x-\mu}{\sigma} \cdot \gamma + \beta} \right ) \cdot (\sigma W_\gamma + B_\gamma) + (\mu W_\beta + B_\beta).
\end{equation}

Compared to BatchNorm~\cite{ioffe2015batch}, LayerNorm is not a cheap operation. 
It needs to compute the mean and standard deviation during inference instead of using the running estimates tracked on the training set.
Therefore, we remove the normalization layer before the MLP, as we find that this hardly worsens the method's performance.

\subsection{Nonlinear Activation Function with Simple Inversal}

\begin{figure}[t]
    \centering
    \includegraphics[width=0.9\columnwidth]{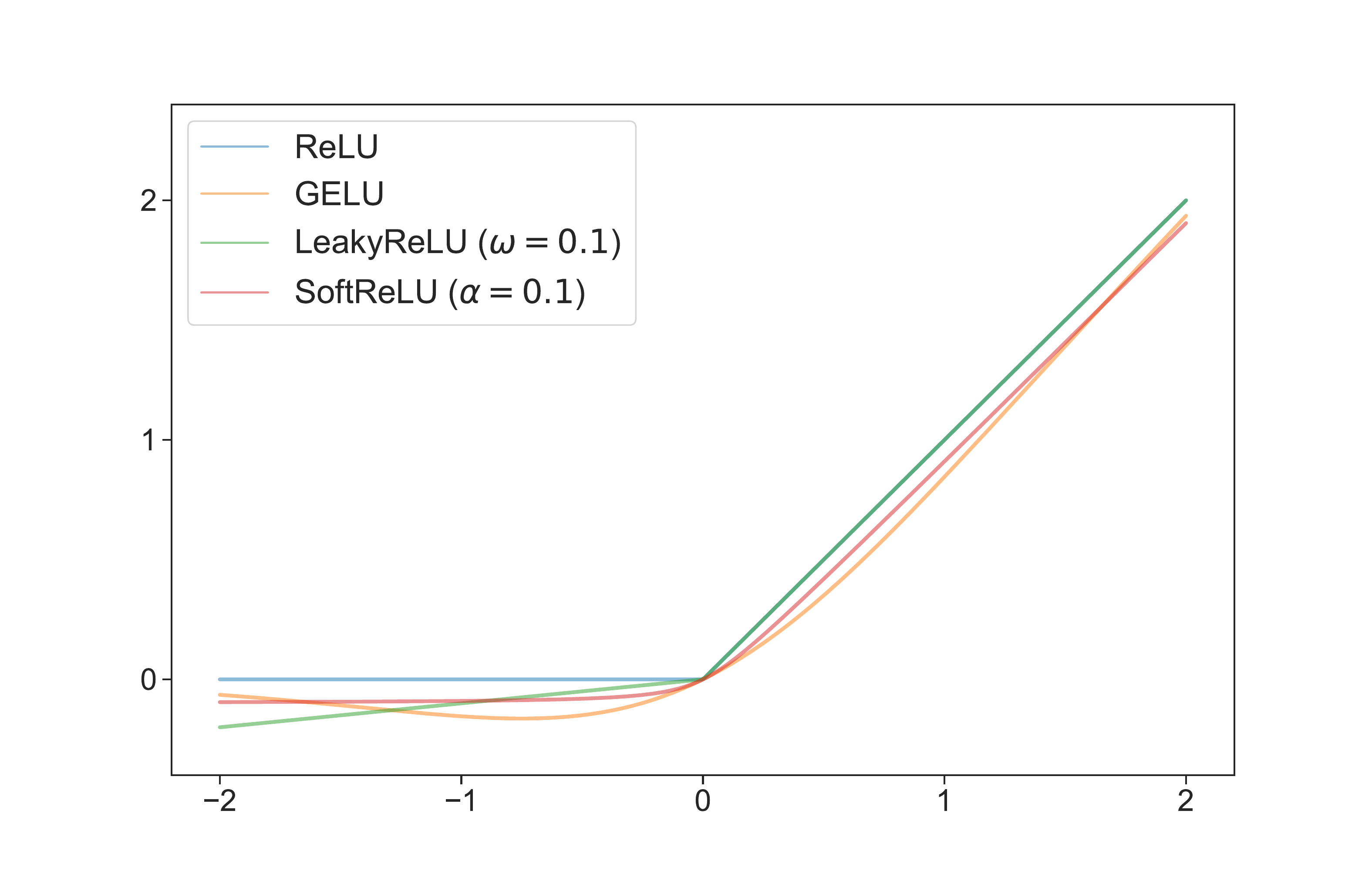}
    \caption{
        The ReLU, GELU, LeakyReLU ($\omega=0.1$), and SoftReLU ($\alpha=0.1$).
    }
    \label{fig:acts}
\end{figure}

GELU performs better than ReLU in high-level tasks~\cite{ramachandran2017searching,tan2019efficientnet,liu2022convnet}.
However, GELU is much less used than ReLU~\cite{nair2010rectified} and LeakyReLU~\cite{maas2013rectifier} in low-level vision tasks.
Although some recent Transformer-based image processing networks inherit GELU~\cite{liang2021swinir,wang2022uformer}, in our experiments, ReLU and LeakyReLU still perform better than GELU in image dehazing.
We believe that GELU does not work in the image dehazing task because it is not easily inverted.
If we consider GELU as an image filter, it causes the gradient reversal problem because of its non-monotonicity.
Unlike high-level vision tasks, the feature maps in image dehazing would be decoded into images, resulting in the reversal gradients introduced by GELU to react in the output image.

Comparing GELU and ReLU, another reason for GELU's inferior performance is its stronger nonlinearity since it is more complicated than piece-wise linear functions.
Hence we propose SoftReLU, which is a simple smooth approximation to the ReLU as an excess between GELU and ReLU:
\begin{equation}
    \text{SoftReLU}(x) = \frac{{x + \sqrt {{x^2} + {\alpha^2}}  - \alpha}}{2}.
\end{equation}
where $\alpha$ is a shape parameter.
In particular, SoftReLU is equivalent to ReLU when we set $\alpha=0$.
To mimic GELU, we set $\alpha=0.1$ in our experiments.

Fig.~\ref{fig:acts} illustrates a comparison of the SoftReLU with other nonlinear activation functions.
We perform ablation studies on activation functions and find that LeakyReLU performs similarly to the ReLU, better than SoftReLU and GELU, while SoftReLU is better than GELU.
Therefore, we believe that the nonlinear activation function's invertibility is essential for image dehazing networks.

\subsection{Shifted Window Partitioning with Reflection Padding}

Swin Transformer uses cyclic shift with masked MHSA to implement efficient batch computation for shifted window partitioning.
Because of the mask, the window size at the edge of the image is smaller than the set window size.
For high-level vision tasks, the object of the image is often in the center of the image, making the edge pixels of the image contribute little to the result.
For image dehazing, image edges are as important as image centers.
A small window size leads to a smaller number of tokens in the window, which biases the network's training.
We consider that the network's performance can be improved by keeping the window size of the image edges the same as the set window size.

\begin{figure}[t]
    \centering
    \includegraphics[width=1.0\columnwidth]{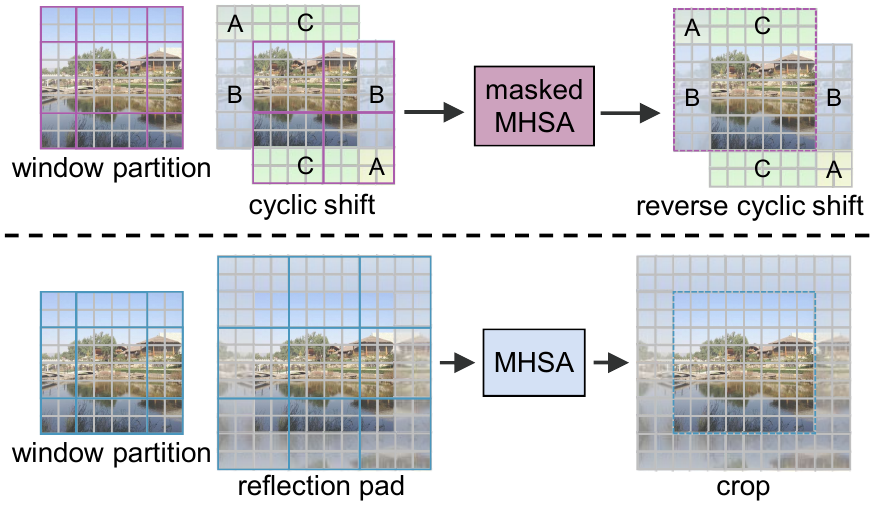}
    \caption{
        Comparison of our proposed reflection padding scheme with cyclic shift scheme for shifted window partitioning. 
        The actual percentage of the edge area is much smaller than the illustration.
    }
    \label{fig:padding}
\end{figure}

To avoid introducing unreasonable inter-patch interactions, we propose to use reflection padding to achieve efficient batch computation for shifted window partitioning, as Fig.~\ref{fig:acts} illustrated.
Swin Transformer's original paper mentions how to use padding to implement batch computation.
However, its proposed padding-based scheme is equivalent to the cyclic shift since the masked MHSA will still be employed.
Unlike Swin Transformer, we use reflection padding and do not perform masking.
The drawback of this method is that it does introduce additional computational costs compared to the cyclic shift. 
Fortunately, image dehazing networks tend to process much larger images than image patches at training time. 
When the image size becomes larger, the percentage of edge regions will become smaller.

\subsection{W-MHSA with Parallel Convolution}

We consider that multiplying MHSA is a low-pass filtering, a similar conclusion was presented in a very recent work~\cite{park2022vision}.
Although the spatial information aggregation weight of MHSA is dynamic, the weight is always positive, making it work like smoothing.
As a counterpart to MHSA's spatial information aggregation style, we perform additional convolution on $V$.
Thus the spatial information aggregation scheme is
\begin{equation}
    \begin{split}
    \text {\small Aggregation}(Q, K, V) & =\operatorname{Softmax}\left({Q K^{T}} / \sqrt{d}+B\right) V \\
    & + \operatorname{Conv}(\hat{V}),
    \end{split}
\end{equation}
where $\hat{V} \in \mathbb{R} ^ {b \times h \times w \times c}$ denotes the $V$ before window partitioning, and $\operatorname{Conv}(\cdot)$ can be either DWConv or a ConvBlock (Conv-ReLU-Conv).
In other words, we still use the attention mechanism to aggregate information within the window, but also use convolution to aggregate information in the neighborhood without considering window partitioning.
Furthermore, we discard the MHSA in some blocks, especially in the encoder's shallow stages and the decoder, and the revised block is shown in Fig.~\ref{fig:arch}.
The components illustrated with dashed boxes in the DehazeFormer block indicate they are optional.
Specifically, some blocks do not contain MHSA and RescaleNorm, and reflection padding and cropping are used only when shifted window partitioning is required.

Note that a similar idea was proposed in CSwin Transformer~\cite{dong2021cswin}, but we use the convolution to extract high frequency information instead of acting as a positional embedding.
In contrast to CSwin Transformer, we use reflection padding instead of zero padding because we do not need it to encode position information implicitly~\cite{islam2020much}.
Most importantly, DehazeFormer's convolutional layer is performed on $\hat{V}$ before window partitioning, thus it provides the capability to aggregate information between windows.

\subsection{Implementation Details}

\begin{table*}[!t]
\centering
\caption{
    Detailed architecture specifications.
}
\label{tab:arch}
\begin{center}
    \renewcommand\arraystretch{1.25}
    {
    \begin{tabular}{|ccccccc|}
    \hline
    & Num. of Blocks & Embedding Dims & MLP Ratio & Attention Ratio & Num. of Heads & Conv Type \\
    \hline
    \hline
    DehazeFormer-T & [\enspace 4, \enspace 4, \enspace 4, \enspace 2, \enspace 2] & [24, 48, \enspace 96, 48, 24] & [2, 4, 4, 2, 2] & [1/4, 1/2, 3/4, 0, 0] & [2, 4, 6, 1, 1] & DWConv \\
    DehazeFormer-S & [\enspace 8, \enspace 8, \enspace 8, \enspace 4, \enspace 4] & [24, 48, \enspace 96, 48, 24] & [2, 4, 4, 2, 2] & [1/4, 1/2, 3/4, 0, 0] & [2, 4, 6, 1, 1] & DWConv \\
    DehazeFormer-B & [16, 16, 16, \enspace 8, \enspace 8] & [24, 48, \enspace 96, 48, 24] & [2, 4, 4, 2, 2] & [1/4, 1/2, 3/4, 0, 0] & [2, 4, 6, 1, 1] & DWConv \\
    DehazeFormer-M & [12, 12, 12, \enspace 6, \enspace 6] & [24, 48, \enspace 96, 48, 24] & [2, 4, 4, 2, 2] & [1/4, 1/2, 3/4, 0, 0] & [2, 4, 6, 1, 1] & ConvBlock \\
    DehazeFormer-L & [16, 16, 16, 12, 12] & [48, 96, 192, 96, 48] & [2, 4, 4, 2, 2] & [1/4, 1/2, 3/4, 0, 0] & [2, 4, 6, 1, 1] & ConvBlock \\
    \hline
    \end{tabular}
    }
\end{center}
\end{table*}

We provide five DehazeFormer's variants (-T, -S, -B, -M, and -L for tiny, small, basic, middle, and large, respectively).
TABLE~\ref{tab:arch} lists the detailed configurations of these variants.
The attention ratio here indicates the percentage of blocks containing MHSA, and we place the blocks containing MHSA at the end of each stage.
For the three small models (-T, -S, -B), we use DWConv ($K = 5$) as the parallel convolutions.
Because DWConv is an operation with low computational cost but high memory access cost~\cite{ma2018shufflenet}, we use ConvBlock ($K = 3$) for two large models (-M, -L).

When training, images are randomly cropped to $256 \times 256$ patches.
We set different mini-batch sizes for training different variants, \emph{i.e.}, $\{32,16,16,16,8\}$ for \{-T, -S, -B, -M, -L\}.
Referring to the linear scaling rule~\cite{goyal2017accurate}, we set the initial learning rate to $\{4,2,2,2,1\}  \times 10^{-4}$ for \{-T, -S, -B, -M, -L\}.
We use AdamW optimizer~\cite{loshchilov2017decoupled} with the cosine annealing strategy~\cite{loshchilov2016sgdr} to train the models, where the learning rate gradually decreases from the initial learning rate to $\{4,2,2,2,1\} \times 10^{-6}$.

\section{RS-Haze dataset}

The RESIDE dataset is a large-scale homogeneous image dehazing dataset that advances the image dehazing.
However, evaluating the method's capability of non-homogeneous image dehazing still relies on some small, unrealistic datasets~\cite{ancuti2020nh}, which use a haze machine to generate the non-homogeneous haze that would hardly exist.
In contrast, remote sensing image dehazing is a practical non-homogeneous image dehazing task because the haze in remote sensing images is highly non-homogeneous.
Therefore, we propose a new synthetic remote sensing image dehazing dataset named RS-Haze.
Comparing to some remote sensing image dehazing datasets~\cite{huang2020single,li2020coarse,mehta2021domain,darbaghshahi2021cloud}, our proposed dataset is more realistic and larger scale.

\subsection{Haze Synthesis Formulation}
For generating remote sensing hazy images, researchers generally set $d(x)$ to $d_0$ since the remote sensing imaging system has a fixed imaging distance. 
However, $d(x)$ is not the imaging distance but the thickness of the medium that scatters the light.
Further, the haze medium in remote sensing images is non-homogeneous, making $d(x)$ vary spatially but consistent over all channels.
Besides, the transmission map $t(x)$ is correlated with wavelength and haze conditions. 
Inspired by prior works~\cite{mccartney1976optics,nayar1999vision}, we model the scattering coefficient as
\begin{equation}
    \beta(\lambda,\gamma(x)) = c_0\lambda^{-\gamma(x)},
    \label{eq:gamma}
\end{equation}
where $c_0$ is a constant, $\lambda$ is the channel's center wavelength, and the exponent $\gamma(x)$ corresponds to the region-wise haze conditions.
Combining Eq.(\ref{eq:depth}) and Eq.(\ref{eq:gamma}), we can derive
\begin{equation}
    t(x) = e^{-\beta(\lambda,\gamma(x)) d(x)} .
    \label{eq:t2}
\end{equation}
Then the relationship of the transmission map between channel $i$ and channel $j$ can be expressed as
\begin{equation}
    \ln t_{i}(x) / \ln t_{j}(x) = \beta_{i}\left(\lambda_{i}, \gamma(x)\right) / \beta_{j}\left(\lambda_{j}, \gamma(x)\right),
    \label{eq:ln2}
\end{equation}
where $t_{\{i,j\}}(x), \beta_{\{i,j\}}, \lambda_{\{i,j\}}$ are the transmission map, scattering coefficient and center wavelength of channel $\{i,j\}$, respectively. 
If we take channel $1$ as the reference channel, and the transmission map $t_j (x)$ can be obtained via
\begin{equation}
    t_j(x) = {t_1(x)}^{\left(\frac{\lambda_1}{\lambda_j}\right)^{\gamma(x)}} ,
    \label{eq:tj}
\end{equation}
The final haze imaging model can be formulated as
\begin{equation}
   I_{j}(x)=J_{j}(x) t_j(x) + A_{j}\left(1-t_j(x)\right).
   \label{eq:asm2}
\end{equation}
Here we can collect clean images $J$ and set ${\lambda}_j$ to the center wavelength of the corresponding channel. 
Thus the problem lies in how to obtain $t_1(x)$, $A_j$ and $\gamma(x)$.

\subsection{Synthesis Pipeline}

We first consider how to extract the transmission map $t_1(x)$ from the real hazy images.
The reflectance of the cirrus channel (channel 9) can characterize the spatially non-homogeneous properties of the natural haze~\cite{guo2020rsdehazenet}, so we use it to generate the transmission map $t_1(x)$ as
\begin{equation}
t_{1}(x) = 1 - {\omega} {\rho}_9(x),
\end{equation}
where ${\rho}_9(x)$ is the reflectance of the cirrus channel of the real hazy image, and $\omega$ is a hyperparameter corresponding to the haze density.
We find a large dark level in the cirrus channel, making the pixels over 5000 even in the haze-free image. 
Thus we apply a linear stretch of 0.1\% to the cirrus channel to remove the dark level.
If we do not remove the dark level, then the maximum value of $t_1(x)$ is always smaller than 1, which is equivalent to an additional homogenous haze.

\begin{figure}[t]
    \centering
    \includegraphics[width=1.0\columnwidth]{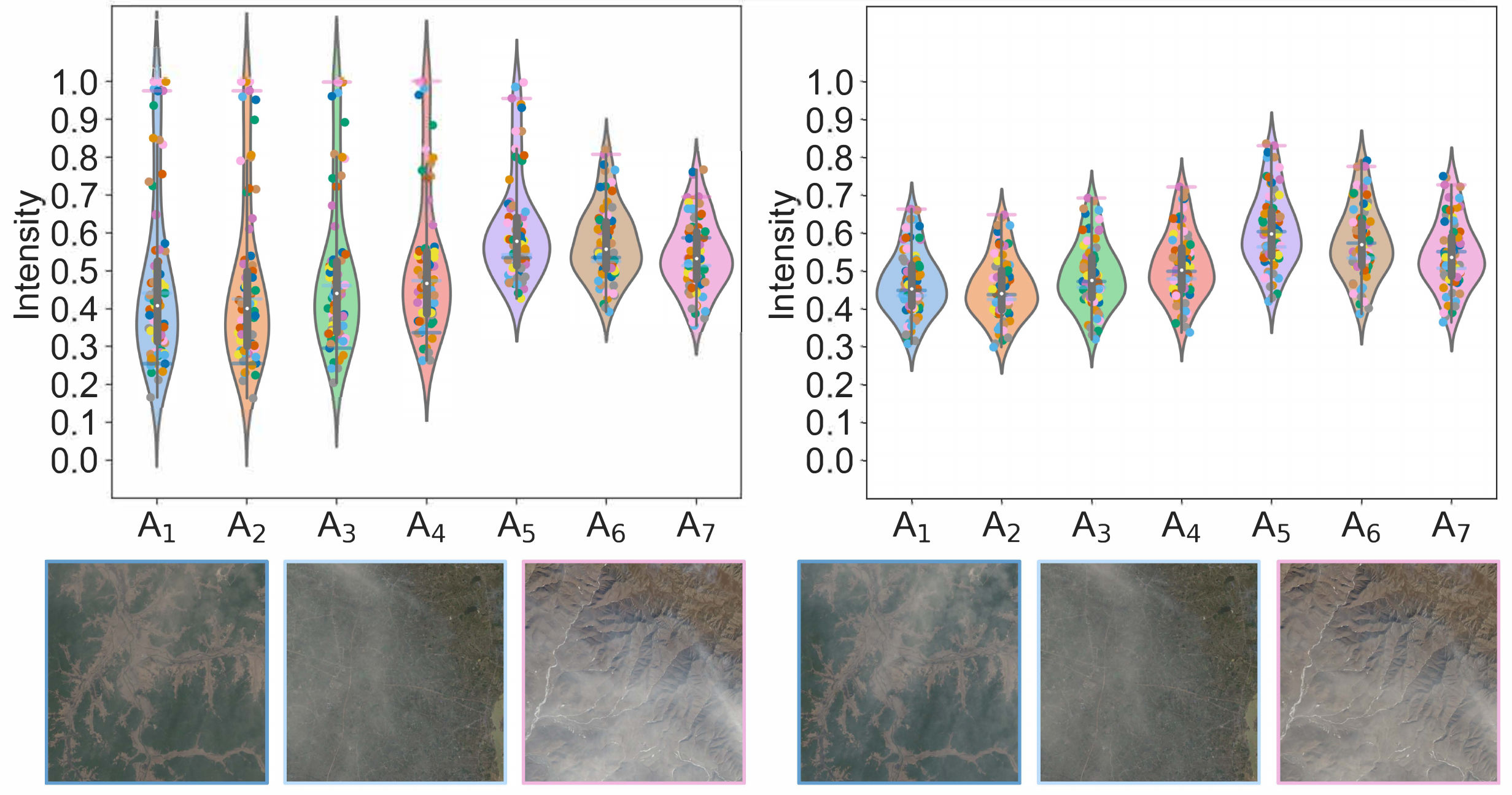}
    \caption{
        Correction of the estimated atmospheric light. 
        The top is atmospheric lights before and after correction, and the bottom is three synthesis samples.
    }
    \label{fig:cal}
\end{figure}

After that, we need to estimate the atmospheric light of the scene from the haze-free images.
To this end, we regard the mean value of each channel's brightest 0.01\% pixels as the atmospheric light\cite{guo2020rsdehazenet}.
However, there are still many cases of inaccurate estimation. 
Since the atmospheric light of each channel is correlated with each other, an additional constraint can be introduced to correct for the incorrectly estimated atmospheric light.
Assume that the mean value of the estimated atmospheric light of all remote sensing images in channel $i$ is $\overline{A_{i}}$.
We set the reference values $\overline{A_{r}} = (\overline{A_{6}} +\overline{A_{7}}) / 2$ and $A_{r} = (A_{6} + A_{7}) / 2$, and obtain the corrected atmospheric light of channel $i$ by $A'_{i} = A_{r} \cdot \overline{A_{i}}/\overline{A_{r}}$.
Fig.~\ref{fig:cal} shows how the correction refines the atmospheric light.

Finally, we need to obtain the $\gamma(x)$.
Because the particle properties of haze can vary depending on the haze density~\cite{chavez1988improved}, the exponent $\gamma(x)$ should be modeled as a function of haze density.
As shown in the Table~\ref{tab:rho_gamma}, we model the exponent $\gamma(x)$ related to the haze reflectance ${\rho}$. 
We use ${\omega} {\rho}_9(x)$ as the haze reflectance and fit the relationship between $\gamma(x)$ and ${\omega} {\rho}_9(x)$ with a cubic curve, which can be formulated as
\begin{equation}
\gamma(x) = a_3({{\omega} {\rho}_9(x)})^3+a_2({{\omega} {\rho}_9(x)})^2+a_1({{\omega} {\rho}_9(x)})+a_0,
\end{equation}
where $a_0=6.537$, $a_1=-27.465$, $a_2=41.224$, and $a_3=-21.547$. 
Note that we clip $\gamma(x)$ to [0, 4] to avoid outliers.

Now we can use Eq.(\ref{eq:asm2}) to synthesize the dataset.
However, we found that the network trained with this dataset works well in the dense haze region of the synthetic image, but performs poorly on the dense haze region of the real image.
We consider that, when the haze is dense, it is likely to block all the light from the ground~\cite{huang2020single}. 
According to the haze imaging model Eq.(\ref{eq:asm2}), even when $t_j(x)$ is small and the synthesized haze is dense, there still exists information residuals from haze-free channel $J_j(x)$.
To this end, we revise Eq.(\ref{eq:asm2}) as
\begin{equation}
    I_{j}(x)=J_{j}(x) {t_{j}(x)}' + A_{j}\left(1-{t_{j}(x)}\right) ,
    \label{eq:asm4}
\end{equation}
where $t_j(x)'=1 - \xi (1-t_j(x))$, and we also clip $t_j(x)'$ to [0, 1] to avoid outliers.
Here $t_j(x)$ is consistent with Eq.(\ref{eq:tj}), but we introduce a decay factor $\xi=1.25$ to attenuate the information of $J_j(x)$.
When the haze reaches a certain concentration, the synthesized hazy image completely lose the information of the hazy-free image in that region. 

\begin{table}[!t]
    \centering
    \renewcommand\arraystretch{1.25}
    \begin{center}
    \caption{Atmospheric Relative Scattering Models}
    \label{tab:rho_gamma}
    \begin{tabular}{|ccc|}
    \hline
    Reflectance~$\rho$ & Condition & Exponent $\gamma$ \\
    \hline
    \hline
    $0.000 < \rho \leq 0.215$ & very clear & 4.0 \\
    $0.215 < \rho \leq 0.294$ & clear & 2.0 \\
    $0.294 < \rho \leq 0.373$ & moderate & 1.0 \\
    $0.373 < \rho \leq 0.451$ & hazy & 0.7 \\
    $0.451 < \rho \leq 0.529$ & very hazy & 0.5 \\
    $0.529 < \rho < 1.000$ & cloudy & 0.0 \\
    \hline
    \end{tabular}
    \end{center}
\end{table}

\subsection{Dataset Details}

\begin{table}[!t]
    \centering
    \renewcommand\arraystretch{1.25}
    \begin{center}
    \caption{Summary of RS-Haze Dataset}
    \label{tab:RS-Haze}
    \begin{tabular}{|ccccc|}
    \hline
    Name & Density & $\omega$ & Train & Test \\
    \hline
    \hline
    RS-Haze-L & Light & 0.100-0.399 & 17100 & 900 \\
    RS-Haze-M & Moderate & 0.400-0.699 & 17100 & 900\\
    RS-Haze-D & Dense & 0.700-0.999 & 17100 & 900\\
    \hline
    RS-Haze-mix & All & 0.100-0.999 & 51300 & 2700 \\
    \hline
    \end{tabular}
    \end{center}
\end{table}

We download the multi-spectral (MS) images from the Landsat-8 Level 1 data product on \href{https://earthexplorer.usgs.gov}{EarthExplorer}.
We selected 76 remote sensing images containing diverse topography with good weather conditions and performed atmospheric correction using the FLAASH module~\cite{cooley2002flaash}.
Meanwhile, 108 cloudy remote sensing images are selected to generate transmission maps using their cirrus channels.
We crop 512$\times$512 image patches from the original remote sensing image using the GDAL library~\cite{warmerdam2008geospatial}.
Finally, we obtained 6000 patches of haze-free MS images containing various terrain and 1500 patches of cirrus channels with a distribution similar to natural haze.
Each haze-free image generates nine synthetic hazed images containing three different haze densities. 
The haze density is controlled by setting the range of $\omega$. 
The values of $\omega$ in each range are obtained by sampling from the truncated Gaussian function.
The summary of RS-Haze is shown in Table~\ref{tab:RS-Haze}. 

\section{Experiments}

\subsection{Experimental setup}

Our experiments are performed on the RESIDE dataset~\cite{li2018benchmarking} and our RS-Haze dataset.
The RESIDE dataset is one of the most commonly used datasets for image dehazing, and it contains three versions: RESIDE-V0, RESIDE-Standard, and RESIDE-$\beta$.
It contains several subsets, of which the most commonly used are: indoor training set (ITS), outdoor training set (OTS), and synthetic objective testing set (SOTS).
We found that the existing works use different experimental setups and can be divided into two main categories: training on a combination of the ITS and the OTS and testing on the SOTS; training on the ITS and the OTS separately and testing on the indoor and outdoor scenes of the SOTS separately.
For proving the effectiveness of DehazeFormer, we perform experiments on both setups, which we name RESIDE-Full and RESIDE-6K, respectively.
We do not train large models under each experimental setup since the small models are good enough.

\subsubsection{RESIDE-Full} 
Models are trained and tested on the indoor and outdoor scenes separately.
Following FFA-Net~\cite{qin2020ffa}, we use the full ITS (13,990 image pairs from RESIDE-Standard) and OTS (313,950 image pairs from RESIDE-V0) to train indoor models and outdoor models and test them on indoor scenes (500 image pairs) and outdoor scenes (500 image pairs) of the SOTS, respectively.
In this experimental setup, all models are trained using their original training strategies, and we replicate the best results reported in the previous works.
We train DehazeFormers on ITS for 300 epochs and on OTS for 30 epochs.
Note that a few images in the outdoor subset are smaller than the configured patch size, so we discard these images during training.
Besides, because the upper part of the outdoor image is often sky, we use only horizontal flipping for data augmentation.

\subsubsection{RESIDE-6K} 
Models are trained and tested on the mixed dataset.
We use an experimental setup from DA~\cite{qin2020ffa}, which differs significantly from the RESIDE-Full.
Its training set contains 3,000 ITS image pairs and 3,000 OTS image pairs, and all images are resized to $400 \times 400$. 
Its testing set mixes indoor and outdoor image pairs to form a test set of 1,000 image pairs without resizing, here called SOTS-mix.
In this experimental setup, we retrain all models using $L_1$ loss on the RESIDE-6K training set for 1,000 epochs, and the learning rate is adjusted according to the model's mini-batch size.
For some methods that estimate $t(x)$, we adapt them to predict the output image.
Thus we can compare the architectures' performance, regardless of the impact of the training strategy.

\subsubsection{RS-Haze} 
Models are trained on the RS-Haze-mix.
For the default experimental setup, we use 8-bit gamma-corrected RGB images for training and testing.
We train all models using $L_1$ loss for 150 epochs, and other settings are the same as RESIDE-6K.
For MS image dehazing, we use 16-bit linear images for training and testing.
It aims to analyze the properties of MS and RGB images for image dehazing.
Note that we compute PSNR and SSIM on the gamma-corrected RGB images when testing.

\subsubsection{Overhead}
We use the number of parameters (\#Param) and multiply-accumulate operations (MACs) to measure the overhead.
MACs are measured on $256\times256$ images.

\subsection{Quantitative Comparison}

\begin{table*}[t]
  \centering
  \caption{
    Quantitative comparison of various dehazing methods trained on the RESIDE datasets.
  }
  \label{tab:quantitative}
  \begin{center}
    \renewcommand\arraystretch{1.25}
    {
      \begin{tabular}{|c|cc|cc|cc|cc|cc|}
        \hline
        \multirow{3}*{Methods}&\multicolumn{2}{c|}{ITS}&\multicolumn{2}{c|}{OTS}&\multicolumn{2}{c|}{RESIDE-6K} &\multicolumn{2}{c|}{RS-Haze} & \multicolumn{2}{c|}{\multirow{2}{*}{Overhead}}\\
        \cline{2-9}
         &\multicolumn{2}{c|}{SOTS-indoor}&\multicolumn{2}{c|}{SOTS-outdoor} &\multicolumn{2}{c|}{SOTS-mix} &\multicolumn{2}{c|}{RS-Haze-mix} & \multicolumn{2}{c|}{}\\
        \cline{2-11}
                                                            & PSNR  & SSIM  & PSNR  & SSIM   & PSNR  & SSIM  & PSNR  & SSIM  & \#Param & MACs \\
        \hline
        \hline
        (TPAMI'10) DCP~\cite{he2010single}                  & 16.62 & 0.818 & 19.13 & 0.815 & 17.88 & 0.816 & 17.86 & 0.734 & -       & -      \\
        (TIP'16) DehazeNet~\cite{cai2016dehazenet}          & 19.82 & 0.821 & 24.75 & 0.927 & 21.02 & 0.870 & 23.16 & 0.816 & 0.009M & 0.581G \\
        (ECCV'16) MSCNN~\cite{ren2016single}                & 19.84 & 0.833 & 22.06 & 0.908 & 20.31 & 0.863 & 22.80 & 0.823 & 0.008M & 0.525G \\
        (ICCV'17) AOD-Net~\cite{li2017aod}                  & 20.51 & 0.816 & 24.14 & 0.920 & 20.27 & 0.855 & 24.90 & 0.830 & \underline{0.002M} & \underline{0.115G} \\
        (CVPR'18) GFN~\cite{ren2018gated}                   & 22.30 & 0.880 & 21.55 & 0.844 & 23.52 & 0.905 & 29.24 & 0.910 & 0.499M & 14.94G \\
        (WACV'19) GCANet~\cite{chen2019gated}               & 30.23 & 0.980 & -     & -     & 25.09 & 0.923 & 34.41 & 0.949 & 0.702M & 18.41G \\
        (ICCV'19) GridDehazeNet~\cite{liu2019griddehazenet} & 32.16 & 0.984 & 30.86 & 0.982 & 25.86 & 0.944 & 36.40 & 0.960 & 0.956M & 21.49G \\
        (CVPR'20) MSBDN~\cite{dong2020multi}                & 33.67 & 0.985 & 33.48 & 0.982 & 28.56 & 0.966 & 38.57 & 0.965 & 31.35M & 41.54G \\
        (ECCV'20) PFDN~\cite{dong2020physics}               & 32.68 & 0.976 & -     & -     & 28.15 & 0.962 & 36.04 & 0.955 & 11.27M & 50.46G \\
        (AAAI'20) FFA-Net~\cite{qin2020ffa}                 & 36.39 & 0.989 & \underline{33.57} & \underline{0.984} & \underline{29.96} & \underline{0.973} & \underline{39.39} & \underline{0.969} & 4.456M & 287.8G \\
        (CVPR'21) AECR-Net~\cite{wu2021contrastive}         & \underline{37.17} & \underline{0.990} & -     & -     & 28.52 & 0.964 & 35.69 & 0.959 & 2.611M & 52.20G \\
        \gr (ours) DehazeFormer-T                           & 35.15 & 0.989 & \textbf{33.71} & 0.982 & \textbf{30.36} & 0.973 & 39.11 & 0.968 & 0.686M & 6.658G \\
        \gr (ours) DehazeFormer-S                           & 36.82 & \textbf{0.992} & \textbf{34.36} & 0.983 & \textbf{30.62} & \textbf{0.976} & \textbf{39.57} & \textbf{0.970} & 1.283M & 13.13G \\
        \gr (ours) DehazeFormer-B                           & \textbf{37.84} & \textbf{0.994} & \textbf{34.95} & 0.984 & \textbf{31.45} & \textbf{0.980} & \textbf{39.87} & \textbf{0.971} & 2.514M & 25.79G \\
        \gr (ours) DehazeFormer-M                           & \textbf{38.46} & \textbf{0.994} & \textbf{34.29} & 0.983 & \textbf{30.89} & \textbf{0.977} & \textbf{39.71} & \textbf{0.971} & 4.634M & 48.64G \\
        \gr (ours) DehazeFormer-L                           & \textbf{40.05} & \textbf{0.996} & -     & -     & -     & -     & -& -& 25.44M & 279.7G \\
        \hline
      \end{tabular}
    }
  \end{center}
\end{table*}

We quantitatively compare the performance of DehazeFormers and baselines, and the results are shown in TABLE~\ref{tab:quantitative}.
Here we underline the best results in baselines and bold the results where DehazeFormers exceed them.
Overall, our proposed DehazeFormers outperformed these baselines.
We argue that the RESIDE-Full indoor set mainly measures the model's capability to handle high-frequency information, and the outdoor set mainly measures the convergence speed of the model.
RESIDE-6K measures the stability of the model and the capability to extract low-frequency information.
RS-Haze measures the network's capability to extract semantic features.
Notably, DehazeFormer-B sometimes outperforms DehazeFormer-M, indicating that the attention mechanism is more critical than convolution in these experimental setups.

\subsubsection{RESIDE-Full}
Training on ITS and testing on SOTS indoor set should be the most widely used experimental setup.
Comparing the baseline methods, FFA-Net and AECR-Net are far superior to the previous or contemporaneous methods. 
The former mainly relies on the large network, and the latter may also benefit from the proposed contrastive loss function.
However, our proposed DehazeFormer-B surpasses all baseline methods in terms of PSNR and SSIM.
Further, the PSNR of DehazeFormer-L exceeds 40 dB. 
To the best of our knowledge, this is the first method with the PSNR exceeding 40 dB on the SOTS indoor set, dramatically surpassing previous work.
Finally, all variants of DehazeFormer work well, and we believe it is a scalable method.
Unfortunately, some baselines do not report results on SOTS outdoor set.
Because the training set of outdoor scenes consists of more than 300,000 sample pairs, DehazeFormers and baselines may not have converged. 
We believe that there is still much scope to improve the performance on SOTS outdoor set, and the current results reflect more the network's convergence speed.
In particular, DehazeFormer-M is inferior to DehazeFormer-S on the outdoor set, probably because more nonlinear activation functions slow down the training.
We remind that the results of baselines on RESIDE-Full are replicated from previous works, and some of them can achieve higher performance using our codebase.

\subsubsection{RESIDE-6K}
We found that the performance of all CNN-based networks under the RESIDE-6K experimental setup is worse than that of DehazeFormers, which we suppose is due to the different image resolutions of the testing and training sets.
Because the images of the training set are resized, its high-frequency information distribution is not consistent with the images of the testing set.
As we argued, the convolutional layer is good at filtering high-frequency information, while the attention mechanism is good at filtering low-frequency information, making DehazeFormers perform better.
We believe this property of the attention mechanism is important for image dehazing because it is not practical to collect dehazing datasets for each resolution setting.

\subsubsection{RS-Haze}
Compared with other experimental setups, the methods have higher PSNR on RS-Haze but lower SSIM.
The scenes of remote sensing images are more monotonous than natural scenes. 
Thus, it is easier for methods to estimate images' latent color and brightness, making the PSNR higher.
In contrast, the haze of RS-Haze is highly non-homogeneous, making the high-frequency information of the image corrupted and the SSIM accordingly lower.
We compare the image dehazing methods on RS-Haze. 
It can be seen that FFA-Net is the best method in baselines, while our small model surpasses it.
It is not only due to the excellent design of DehazeFormer itself but also because the remote sensing images have more similar regions, which are more favorable for self-attention~\cite{mei2020pyramid}.
Furthermore, the comparison of DehazeFormer-S on RGB and MS images is shown in TABLE~\ref{tab:ms}.
As expected, dense haze is more difficult to be removed than light haze.
Besides, the additional information provided by more channels and larger bit depths does improve the performance of the method substantially.

\begin{table}[t]
  \centering
  \caption{
    PSNR / SSIM of DehazeFormer-S on the RGB / MS set.
  }
  \label{tab:ms}
  \begin{center}
    \renewcommand\arraystretch{1.25}
      {
      \begin{tabular}{|c|ccc|c|}
        \hline
        Setting & RS-Haze-L & RS-Haze-M  & RS-Haze-D & RS-Haze-mix  \\
        \hline
        \hline
        RGB & 43.68/0.993 & 39.58/0.979 & 35.46/0.938 & 39.57/0.970  \\
        MS & 55.44/0.999 & 50.75/0.997 & 43.66/0.984 & 49.95/0.993  \\
        \hline
      \end{tabular}
      }
  \end{center}
\end{table}

\subsection{Qualitative Comparison}

\begin{figure*}[!t]
  \centering
  \includegraphics[width=1.0\textwidth]{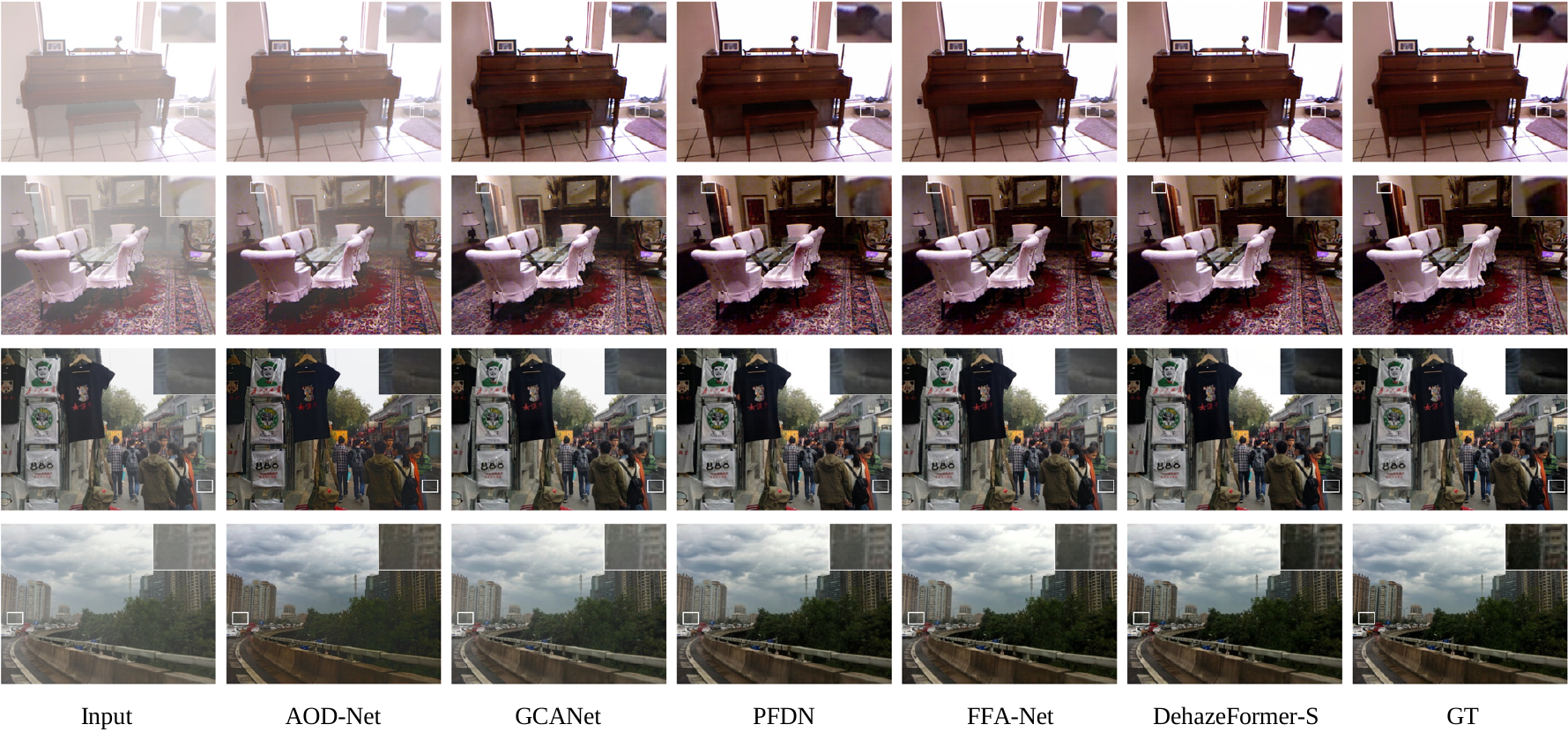}
  \caption{
    Qualitative comparison of image dehazing methods on SOTS mix set, where the first two rows are indoor images, and the last two rows are the outdoor images.
    The first column is the hazy images and the last column is the corresponding ground truth.
  }
  \label{fig:compare1}
\end{figure*}

\begin{figure*}[!t]
  \centering
  \includegraphics[width=1.0\textwidth]{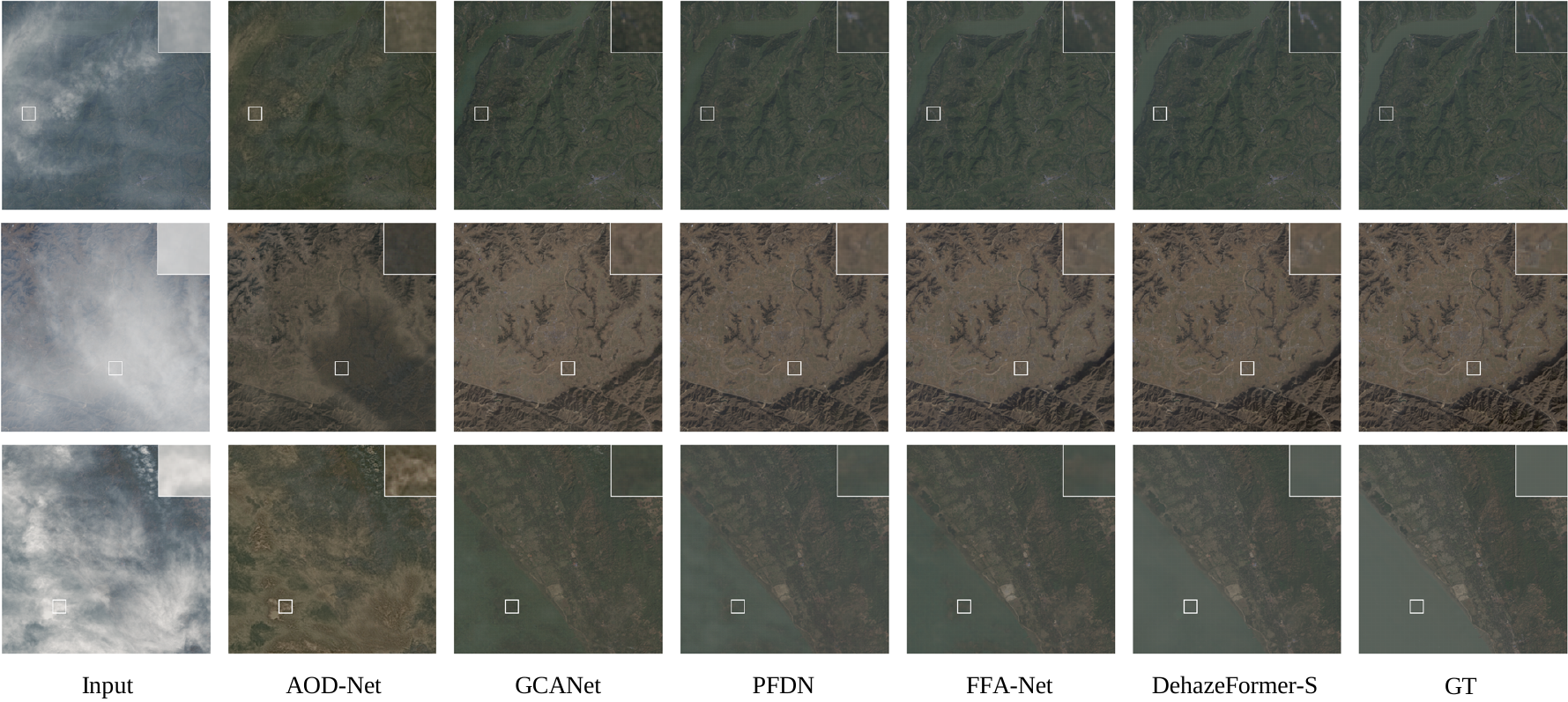}
  \caption{
    Qualitative comparison of image dehazing methods on RS-Haze.
    The first column is the hazy images and the last column is the ground truth.
  }
  \label{fig:rscompare}
\end{figure*}

We also select some samples of the results to analyze the performance of each method qualitatively.
Since we do not retrain the baselines on RESIDE-Full, we only show the test results on RESIDE-6K and RS-Haze.
Fig.~\ref{fig:compare1} and Fig.~\ref{fig:rscompare} illustrate qualitative comparisons of our DehazeFormer-S with some representative learning-based dehazing methods. 

\subsubsection{RESIDE-6K}
We select four samples taken from different scenes in the SOTS mix set to evaluate the network's dehazing performance, including synthetic indoor and outdoor haze. 
AOD-Net and GCANet produce severe color distortions, which make their indoor and outdoor results too dim or too bright.
Though color is restored in most areas of images dehazed by PFDN and FFA-Net, color distortion remains on distant objects and small objects near the edge of images. 
By comparison, the color is recovered correctly through the haze by our DehazeFormer-S, and the results look natural and realistic.
When it comes to the region where haze density varies significantly in some indoor scenes, as shown in the enlarged white boxes in the second row in Fig.~\ref{fig:compare1}, we can observe that almost all the comparative methods fail to remove the haze effectively. 
However, our DehazeFormer-S restores clear images well, keeps texture and color information, and contains the least haze residual.

\subsubsection{RS-Haze}
Three images taken from different scenes with different haze densities in the RS-Haze are selected to evaluate the network's dehazing performance on non-homogeneous haze. 
AOD-Net can barely handle non-homogeneous haze and produces severe artifacts.
GCANet, PFDN, and FFA-Net can remove haze effectively when the haze is thin, as shown in the first two rows in Fig.~\ref{fig:rscompare}, but they are not as good as DehazeFormer-S in color and detail reproduction.
Moreover, DehazeFormer-S can remove dense haze, while all other networks produce apparent artifacts. 
See the water surface area in the third row of Fig.~\ref{fig:rscompare}.

\subsection{Ablation Study}

\begin{table*}[t]
  \centering
  \caption{
    Comparison between DehazeFormer-T and DehazeFormer-A.
  }
  \label{tab:ablation_arch}
  \begin{center}
    \renewcommand\arraystretch{1.25}
    {
    \begin{tabular}{|c|cccc|cccc|}
    \hline
    & Num. of Blocks & MLP Ratio & Attention Ratio & Num. of Heads & PSNR & SSIM & \#Param & MACs \\
    \hline
    \hline
    \gr DehazeFormer-T & [4, 4, 4, 2, 2] & [2, 4, 4, 2, 2] & [1/4, 1/2, 3/4, 0, 0] & [2, 4, 6, 1, 1] & 35.15 & 0.989 & 0.686M & 6.658G\\
    DehazeFormer-A & [2, 2, 2, 2, 2] & [2, 4, 4, 4, 2] & [\hspace{0.3pt} 1 \hspace{0.3pt}, \hspace{0.3pt} 1 \hspace{0.3pt}, \hspace{0.3pt} 1 \hspace{0.3pt}, 1, 1] & [2, 4, 6, 4, 2] & \bt{34.85} & \bt{0.988} & \rt{0.671M} & \bt{7.185G}\\
    \hline
    \end{tabular}
    }
\end{center}
\end{table*}

\begin{table}[t]
  \centering
  \caption{
    Ablation study on normalization layers.
  }
  \label{tab:ablation1}
  \begin{center}
    \renewcommand\arraystretch{1.25}
      {
      \begin{tabular}{|l|cccc|}
        \hline
        Setting & PSNR & SSIM  & \#Param & MACs  \\
        \hline
        \hline
        \gr DehazeFormer-A & 34.85 & 0.988 & 0.671M & 7.185G  \\
        RescaleNorm $\rightarrow$ LayerNorm$^\dag$ & \bt{34.73} & 0.988 &  \rt{0.668M} & \rt{7.175G}  \\
        \hspace{43.5pt} $\rightarrow$ LayerNorm & \bt{34.45} & \bt{0.987} &  \rt{0.668M} & \rt{7.175G}  \\
        $-$ PreNorm for MHSA & \bt{34.17} & \bt{0.986} & \rt{0.668M}  & \rt{7.163G} \\
        $+$ PreNorm for MLP & \rt{34.86} & \bt{0.987} & \bt{0.674M}  & \bt{7.207G} \\
        \hline
      \end{tabular}
      }
  \end{center}
\end{table}

\begin{table}[t]
  \centering
  \caption{
    Ablation study on shifted window partitioning schemes.
  }
  \label{tab:ablation2}
  \begin{center}
    \renewcommand\arraystretch{1.25}
      {
      \begin{tabular}{|l|cccc|}
        \hline
        Setting & PSNR & SSIM  & \#Param & MACs  \\
        \hline
        \hline
        \gr DehazeFormer-A & 34.85 & 0.988 & 0.671M & 7.185G  \\
        $\rightarrow$ Zero-Padding  & \bt{34.00} & \bt{0.986} & 0.671M  & 7.185G \\
        $\rightarrow$ Padding w/ Mask  & \bt{34.54} & 0.988 & 0.671M & 7.185G \\
        $\rightarrow$ Cyclic Shift w/ Mask & \bt{34.54} & 0.988 & 0.671M  & \rt{7.106G} \\
        $\rightarrow$ Cyclic Shift w/o Mask  & \bt{34.34} & 0.988 & 0.671M  & \rt{7.106G} \\
        \hline
      \end{tabular}
      }
  \end{center}
\end{table}

\begin{table}[t]
  \centering
  \caption{
    Ablation study on nonlinear activation functions.
  }
  \label{tab:ablation3}
  \begin{center}
    \renewcommand\arraystretch{1.25}
      {
      \begin{tabular}{|l|cccc|}
        \hline
        Setting & PSNR & SSIM  & \#Param & MACs  \\
        \hline
        \hline
        \gr DehazeFormer-A & 34.85 & 0.988 & 0.671M & 7.185G  \\
        ReLU $\rightarrow$ GELU  & \bt{33.02} & \bt{0.983} & 0.671M  & \bt{7.279G} \\
        \hspace{19.5pt} $\rightarrow$ SoftReLU ($0.1$)  & \bt{33.73} & \bt{0.985} & 0.671M  & \bt{7.229G} \\
        \hspace{19.5pt} $\rightarrow$ LeakyReLU ($0.1$)  & \rt{34.89} & 0.988 & 0.671M  & \bt{7.229G} \\
        \hline
      \end{tabular}
      }
  \end{center}
\end{table}

\begin{table}[t]
  \centering
  \caption{
    Ablation study on parallel conv layers.
  }
  \label{tab:ablation4}
  \begin{center}
    \renewcommand\arraystretch{1.25}
      {
      \begin{tabular}{|l|cccc|}
        \hline
        Setting & PSNR & SSIM  & \#Param & MACs  \\
        \hline
        \hline
        \gr DehazeFormer-A & 34.85 & 0.988 & 0.671M & 7.185G \\
        $-$ Parallel DWConv & \bt{32.90} & \bt{0.983} & \rt{0.653M} & \rt{6.910G} \\
        $\rightarrow$ Parallel DWConv on $X$ & \bt{33.99} & \bt{0.987} & 0.671M & 7.185G \\
        $\rightarrow$ DWConv before MLP & \bt{33.49} & \bt{0.986} & 0.671M & 7.185G \\
        \hline
      \end{tabular}
      }
  \end{center}
\end{table}

\begin{table}[!t]
  \centering
  \caption{
    Ablation study on other components.
  }
  \label{tab:ablation5}
  \begin{center}
    \renewcommand\arraystretch{1.25}
      {
      \begin{tabular}{|l|cccc|}
        \hline
        Setting & PSNR & SSIM  & \#Param & MACs  \\
        \hline
        \hline
        \gr DehazeFormer-A & 34.85 & 0.988 & 0.671M & 7.185G  \\
        Soft Recon. $\rightarrow$ Recon. & \bt{34.50} & \bt{0.987} & \rt{0.671M}  & \rt{7.171G} \\
        SK Fusion $\rightarrow$ Cat Fusion & \bt{34.78} & 0.988 & \bt{0.673M}  & \bt{7.256G} \\
        \hline
      \end{tabular}
      }
  \end{center}
\end{table}

We perform ablation studies on the RESIDE-Full's indoor scene.
However, because not every DehazeFormer block has an MHSA, these blocks degenerate into meaningless linear layers when removing the parallel convolution.
So we build DehazeFormer-A for ablation studies only.
In particular, we set the attention ratio of DehazeFormer-A to 1 and reduce the depth of the network to keep the computational cost and parameters.
TABLE~\ref{tab:ablation_arch} lists the difference between DehazeFormer-T and DehazeFormer-A.
We can see that DehazeFormer-T has better performance than DehazeFormer-A.
In terms of overhead, DehazeFormer-A has fewer parameters but a higher computational cost compared to DehazeFormer-T.
Note that we find that the results of ablation studies on different datasets are not always consistent, \emph{e.g.}, RESIDE-6K prefers DehazeFormer with a high attention ratio compared to the RESIDE-Full indoor set.
We mark the results in \rt{red} if there is an improvement compared to the baseline (DehazeFormer-A) and in \bt{blue} if there is a degradation.

\subsubsection{Normalization layer}
We study normalization layers and their placements on the performance, and the results are shown in TABLE~\ref{tab:ablation1}.
We can see that avoiding the loss of inter-patch relativity and reintroducing the lost statistics does improve the networks' performance.
Besides, the normalization layer is more critical for MHSA than MLP. 
Considering that the normalization layer before MLP has no significant impact on the performance, removing it makes sense since it is not cheap to obtain the standard deviation of the feature maps.
However, the negative impact of LayerNorm is not as evident as expected since the normalization layer showed a severe impact on performance in our early ablation studies on RESIDE-6K.
Thus we plan to explore the relationship between the dataset and the normalization layer in our future work.

\subsubsection{Shifted window partitioning scheme}
We study the schemes of shifted window partitioning, and the results are shown in TABLE~\ref{tab:ablation2}.
Because masked padding and masked cyclic shift are equivalent in terms of spatial information aggregation, we train only a single network.
If we replace the reflection padding with zero padding, the network's performance drops significantly. 
Zero-padding introduces meaningless tokens, and the attention matrix is all-positive, making useless information mixed in.
In contrast, cyclic shift without mask also introduces unreasonable interactions between tokens but has a less negative impact.
Finally, our proposed scheme gives a moderate performance improvement to the network. 
Considering that it only introduces a negligible additional computational cost on $256 \times 256$ images, it is worthwhile.

\subsubsection{Nonlinear activation functions}
We study the difference in the nonlinear activation functions, and the results are shown in TABLE~\ref{tab:ablation3}.
We replace all nonlinear activation functions in the network, including the nonlinear activation functions in MLP and SK fusion layer.
Surprisingly, the nonlinear activation functions dramatically affect the network performance, while our early ablation studies on RESIDE-6K did not show such a huge gap.
The networks using ReLU and LeakyReLU perform about the same because they are both piecewise linear functions that can be easily inverted.
Although the form of SoftReLU is simple, it is not easily inverted, so the networks with it yield significant performance degradation.
Furthermore, GELU is non-monotonic, and it is more difficult to be inverted, making the networks with it perform very poorly.
We argue that it is essential to consider the invertibility of the nonlinear activation function when building the network.

\subsubsection{Parallel conv}
We study to prove the importance of parallel convolution with attention: a) remove the parallel convolution; b) place the convolution parallel with the MHSA, \emph{i.e.}, the input to the convolution is $X$ instead of $V$; c) place the convolution before the MLP~\cite{liu2022convnet}, and the results are shown in TABLE~\ref{tab:ablation4}.
As can be seen, additional convolutional layers in the Transformer block can dramatically improve the network's performance, but their placement is critical.
Inserting DWConv into the FFN brings only minor performance, although the scheme has been widely employed in previous work.
We consider that the transformer works somehow because it separates intra-token and inter-tokens interactions into two steps, while inserting DWConv in FFN would break this property.
DWConv in parallel with attention is better than DWConv in parallel with MHSA.
Although both schemes use DWConv to aggregate spatial information, the former is done in the same feature space as attention, while the latter is done in a different feature space.
DWConv provides static learnable aggregation weights, while attention provides dynamic all-positive aggregation weights.
Thus the convolution parallel to attention does play a complementary role with attention.

\subsubsection{Other components}

We verify the impact of the soft reconstruction module and SK fusion module on the network's performance.
Although SK fusion brings only a minor performance gain, we consider it a good alternative to concatenation fusion, given its lower overhead.
Whereas the soft reconstruction brings more improvement than expected, we believe introducing soft constraints on prior is beneficial.

\section{Conclusion}

This paper introduces various improvements for Swin Transformer applied to image dehazing, and the proposed DehazeFormer achieves superior performance on several datasets.
To summarize, we propose to use RescaleNorm and ReLU to replace the commonly used LayerNorm and GELU to avoid some negative effects that are not important for high-level vision tasks but critical for low-level vision tasks.
To improve the capability of MHSA, we propose a shifted window partitioning scheme based on reflection padding and a spatial information aggregation scheme using convolution in parallel with attention. 
We also propose some minor improvements that are applicable to other networks.
Finally, we collect a large-scale remote sensing image dehazing dataset to evaluate the network's capability to remove highly non-homogeneous haze, and DehazeFormer also achieves an impressive performance.
In the future, we plan to work on more lightweight and more straightforward architectures and extend the architecture to other low-level vision tasks. 
Besides, encoding feature maps on thumbnails and then decoding them on the original image may achieve real-time 4K image dehazing.


{\small
\bibliographystyle{unsrt}
\bibliography{egbib}
}



\end{document}